\documentclass[runningheads]{llncs}
\usepackage{graphicx}
\usepackage{comment}
\usepackage{amsmath,amssymb} 
\usepackage{color}
\usepackage{multirow}
\usepackage{mathtools} 
\usepackage{bbding}
\usepackage{tikz}  
\usepackage{pgfplots}
\usetikzlibrary{fit}
\pgfplotsset{compat=newest}%
\newcommand{\eg}{\textit{e}.\textit{g}.}
\newcommand{\ie}{\textit{i}.\textit{e}.}

\newlength\savewidth
\usepackage[width=122mm,left=12mm,paperwidth=146mm,height=193mm,top=12mm,paperheight=217mm]{geometry}
\usepackage[pagebackref=true,breaklinks=true,letterpaper=true,colorlinks,bookmarks=false]{hyperref}
\usepackage{array}

\begin{document}
\pagestyle{headings}
\mainmatter
\def\ECCVSubNumber{2057}  

\title{SipMask: Spatial Information Preservation for Fast Image and Video Instance Segmentation} 


\titlerunning{SipMask: Spatial Information Preservation for Fast Instance Segmentation}
%
\author{Jiale Cao$^{1}$\thanks{The first two authors contribute equally. $^\dagger$ Y. Pang is corresponding author.}, Rao Muhammad Anwer$^{2,3\star}$, Hisham Cholakkal$^{2,3}$, \\Fahad Shahbaz Khan$^{2,3}$, Yanwei Pang$^{1\dagger}$, Ling Shao$^{2,3}$}
\authorrunning{J. Cao et al.}
%
\institute{$^1$Tianjin University~~~~~$^2$Mohamed bin Zayed University of Artificial Intelligence, UAE\\$^3$Inception Institute of Artificial Intelligence, UAE\\
{\tt\small $^1$\{connor,pyw\}@tju.edu.cn\\}
{\tt\small $^2$\{rao.anwer,hisham.cholakkal,fahad.khan,ling.shao\}@mbzuai.ac.ae}}
\maketitle

\begin{abstract}
Single-stage instance segmentation approaches have recently gained popularity due to their speed and simplicity, but are still lagging behind in accuracy, compared to two-stage methods. We propose a fast single-stage instance segmentation method, called SipMask, that preserves instance-specific spatial information by separating mask prediction of an instance to different sub-regions of a detected bounding-box. Our main contribution is a novel light-weight spatial preservation (SP) module that generates a separate  set of spatial coefficients for each sub-region within a bounding-box, leading to improved mask predictions. It also enables accurate delineation of spatially adjacent instances. Further, we introduce a mask alignment weighting loss and a feature alignment scheme to better correlate mask prediction with object detection. On COCO \texttt{test-dev}, our SipMask outperforms the existing single-stage methods. Compared to the state-of-the-art single-stage TensorMask, SipMask obtains an absolute gain of 1.0\% (mask AP), while providing a four-fold speedup. In terms of real-time capabilities, SipMask outperforms YOLACT with an absolute gain of 3.0\% (mask AP) under similar settings, while operating at comparable speed on a Titan Xp. We also evaluate our SipMask for real-time video instance segmentation, achieving promising results on YouTube-VIS dataset. The source code is available at \url{https://github.com/JialeCao001/SipMask}.

\end{abstract}

\section{Introduction}

Instance segmentation aims to classify each pixel in an image into an object category. Different from semantic segmentation \cite{Long_FCN_CVPR_2015,Chen_Deeplab_PAMI_2017,Cao_TripleNet_CVPR_2019,Pang_SeENet_ICCV_2019,Sun_Mining_ECCV_2020}, instance segmentation also differentiates multiple object instances. Modern instance segmentation methods typically adapt object detection frameworks, where bounding-box detection is first performed, followed by segmentation inside each of detected bounding-boxes. Instance segmentation approaches can generally be divided into two-stage~\cite{He_MaskRCNN_ICCV_2017,Liu_PANet_CVPR_2018,Chen_HTC_CVPR_2019,Huang_MSRCNN_CVPR_2019,Fang_InstaBoost_ICCV_2019} and single-stage~\cite{Pinheiro_ROS_ECCV_2016,Dai_InstanceFCN_ECCV_2016,Xu_ESE_ICCV_2019,Bolya_YOLACT_ICCV_2019,Wang_RDSNet_AAAI_2020,Peng_DeepSnake_CVPR_2020} methods, based on the underlying detection framework. Two-stage methods typically generate  multiple object proposals in the first stage. In the second stage, they  perform feature pooling operations on each proposal, followed by  box regression, classification, and mask prediction. Different from two-stage methods, single-stage approaches do not require proposal generation or pooling operations and employ dense predictions of bounding-boxes and instance masks. Although two-stage methods dominate accuracy, they are generally slow, which restricts their usability in real-time applications.

As discussed above, most single-stage methods are inferior in accuracy, compared to their two-stage counterparts. A notable exception is the single-stage TensorMask~\cite{Chen_TensorMask_ICCV_2019}, which achieves comparable  accuracy to two-stage methods. However, TensorMask achieves this accuracy at the cost of reduced speed. In fact, TensorMask~\cite{Chen_TensorMask_ICCV_2019} is slower than several two-stage methods, including Mask R-CNN~\cite{He_MaskRCNN_ICCV_2017}. Recently, YOLACT~\cite{Bolya_YOLACT_ICCV_2019} has shown to achieve an optimal tradeoff between speed and accuracy. On the COCO benchmark \cite{Lin_COCO_ECCV_2014}, the single-stage YOLACT operates at real-time (33 frames per second), while obtaining competitive accuracy. YOLACT achieves real-time speed mainly by avoiding proposal generation and feature pooling head networks that are commonly employed in two-stage methods. While operating at real-time, YOLACT still lags behind modern two-stage methods (\eg, Mask R-CNN \cite{He_MaskRCNN_ICCV_2017}), in terms of accuracy.

\begin{figure}[t!]
\centering
\includegraphics[width=1.0\linewidth]{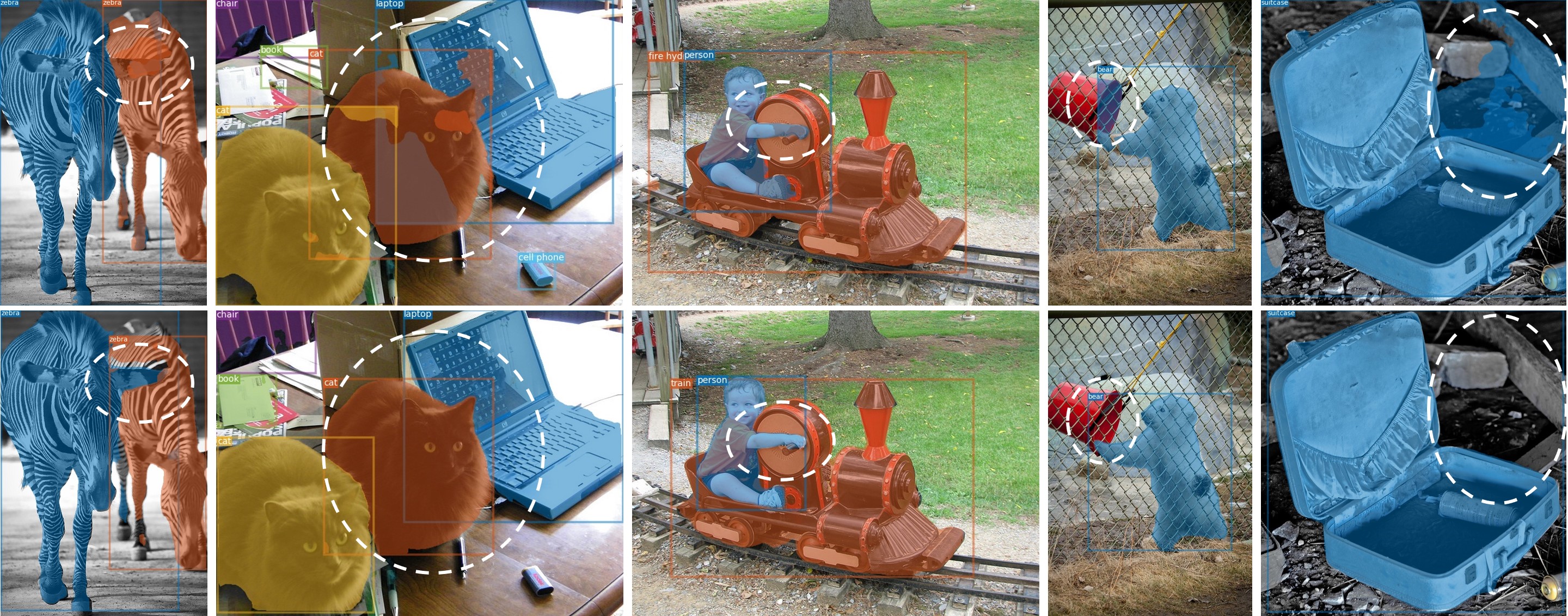} 
\caption{Instance segmentation examples using YOLACT~\cite{Bolya_YOLACT_ICCV_2019} (top) and our approach (bottom). YOLACT struggles to accurately delineate spatially adjacent instances. Our approach with novel spatial coefficients addresses this issue (marked by white dotted region) by preserving spatial information in bounding-box. The spatial coefficients split mask prediction into  multiple  sub-mask  predictions, leading to improved mask quality.}
\label{fig:vis_compare} 
\end{figure}

In this work, we argue that one of the key reasons behind sub-optimal accuracy of YOLACT is the loss of spatial information within an object (bounding-box). We attribute this loss of spatial information due to the utilization of a \textit{single} set of object-aware coefficients to predict the whole mask of an object. As a result, it struggles to accurately delineate spatially adjacent object instances (Fig.~\ref{fig:vis_compare}). To address this issue, we introduce an approach that comprises a novel computationally efficient spatial preservation (SP) module to preserve spatial information in a bounding-box. Our SP module predicts object-aware \textit{spatial} coefficients that splits mask prediction into  multiple sub-mask predictions, thereby enabling improved delineation of spatially adjacent objects (Fig.~\ref{fig:vis_compare}).

\noindent{\textbf{Contributions:}} We propose a fast anchor-free single-stage instance segmentation approach, called SipMask, with the following contributions. 
\begin{itemize}
    \item We propose a novel light-weight spatial preservation (SP) module that preserves the spatial information within a bounding-box. Our SP module generates a separate set of \textit{spatial} coefficients for each bounding-box sub-region, enabling improved delineation of spatially adjacent objects.
     \item We introduce two strategies to better correlate mask prediction with object detection. First, we propose a mask alignment weighting loss that assigns higher weights to the mask prediction errors occurring at accurately detected boxes. Second, a feature alignment scheme is introduced to improve the feature representation for both box classification and spatial coefficients.
     \item Comprehensive experiments are performed on COCO benchmark \cite{Lin_COCO_ECCV_2014}. Our single-scale inference model based on ResNet101-FPN backbone outperforms state-of-the-art single-stage TensorMask~\cite{Chen_TensorMask_ICCV_2019} in terms of \textit{both} mask accuracy (absolute gain of 1.0\% on COCO \texttt{test-dev}) and speed (four-fold speedup). Compared with real-time YOLACT \cite{Bolya_YOLACT_ICCV_2019}, our SipMask provides an absolute gain of 3.0\% on COCO \texttt{test-dev}, while operating at comparable speed.
     \item The proposed SipMask can be extended to single-stage video  instance  segmentation  by  adding a fully-convolutional branch for tracking instances across video frames. On YouTube-VIS dataset~\cite{Yang_VIS_ICCV_2019}, our single-stage approach achieves favourable performance while operating at real-time (30 fps). 
\end{itemize}

\section{Related Work} \label{related_work}
\begin{figure}[t!]
\centering
\includegraphics[width=0.95\linewidth]{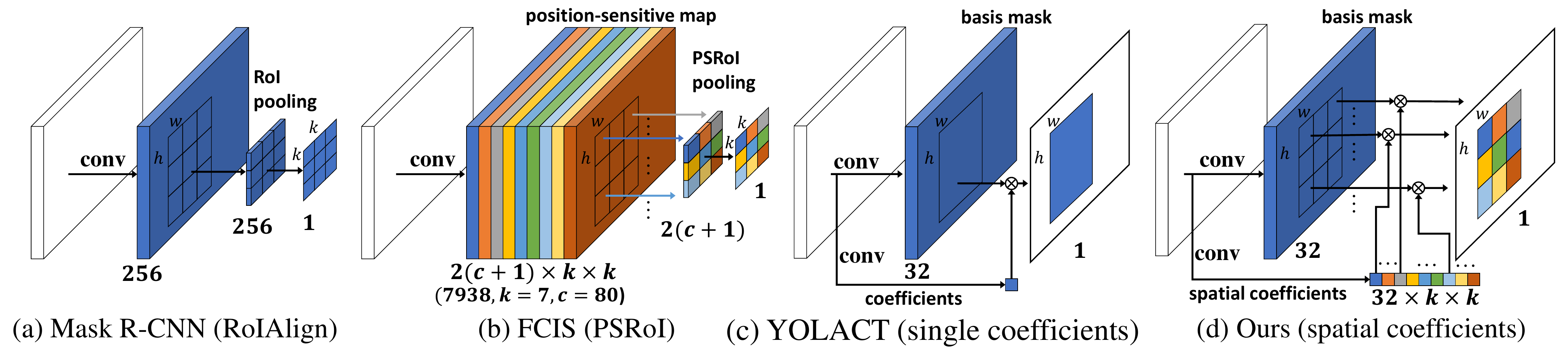} 
\caption{On the left (a and b), feature pooling strategies employed in Mask R-CNN~\cite{He_MaskRCNN_ICCV_2017} and FCIS~\cite{Li_FCIS_CVPR_2017} resize the feature map to a fixed resolution. Instead, both YOLACT~\cite{Bolya_YOLACT_ICCV_2019} (c) and our approach (d) do not utilize any pooling operation and obtain mask prediction by a simple linear combination of basis mask and coefficient. 
Mask R-CNN is computationally expensive ($conv$ and $deconv$ operations after RoIAlign), whereas 
FCIS is memory demanding due to large number of channels in position-sensitive maps. Both YOLACT and our approach reduce the computational and memory complexity. However, YOLACT uses a single set of coefficients for a detected box, thereby ignoring the spatial information within a box. Our approach preserves the spatial information of an instance by using separate set of spatial coefficients for  $k \times k$ sub-regions within a box.}
\label{fig:progress}
\end{figure}

Deep learning has achieved great success in a variety of computer vision tasks \cite{Girshick_RCNN_CVPR_2014,Cholakkal_OCIS_CVPR_2019,Wang_RFHS_ICCV_2019,Pang_MGAN_ICCV_2019,Wu_TFAN_CVPR_2020,Wang_IPNet_CVPR_2020,AR_TIP_2015,Jiang_DAML_TMM_2020,cvpr19uel,arxiv20reidsurvey}. Existing instance segmentation methods either follow bottom-up~\cite{Arnab_PIS_CVPR_2017,Kirillov_InstanceCut_CVPR_2017,Liu_SGN_ICCV_2017,Neven_OSECB_CVPR_2019,Gao_SSAP_ICCV_2019} or top-down~\cite{He_MaskRCNN_ICCV_2017,Liu_PANet_CVPR_2018,Chen_HTC_CVPR_2019,Bolya_YOLACT_ICCV_2019,Peng_DeepSnake_CVPR_2020} paradigms. Modern instance segmentation approaches typically follow top-down paradigm 
where the bounding-boxes are first detected and second segmented. The top-down approaches are divided into two-stage~\cite{He_MaskRCNN_ICCV_2017,Liu_PANet_CVPR_2018,Chen_HTC_CVPR_2019,Huang_MSRCNN_CVPR_2019,Fang_InstaBoost_ICCV_2019} and single-stage~\cite{Dai_InstanceFCN_ECCV_2016,Xu_ESE_ICCV_2019,Bolya_YOLACT_ICCV_2019,Wang_RDSNet_AAAI_2020,Peng_DeepSnake_CVPR_2020} methods.
Among these two-stage methods, Mask R-CNN~\cite{He_MaskRCNN_ICCV_2017} employs a proposal generation network (RPN) and utilizes RoIAlign feature pooling strategy (Fig.~\ref{fig:progress}(a)) to obtain a fixed-sized features of each proposal. The pooled features are used for box detection and mask prediction. A position sensitive feature pooling strategy, PSRoI~\cite{Dai_RFCN_NIPS_2016} (Fig.~\ref{fig:progress}(b)), is proposed in FCIS~\cite{Li_FCIS_CVPR_2017}. PANet~\cite{Liu_PANet_CVPR_2018} proposes an adaptive feature pooling that allows each proposal to access information from multiple layers of FPN. MS R-CNN~\cite{Huang_MSRCNN_CVPR_2019} introduces an additional branch to predict mask quality (mask-IoU). MS R-CNN  performs a mask confidence rescoring without improving mask quality. In contrast, our mask alignment loss  aims to improve mask quality at accurate detections.

Different to two-stage methods, single-stage approaches~\cite{Bolya_YOLACT_ICCV_2019,Dai_InstanceFCN_ECCV_2016,Xu_ESE_ICCV_2019,Wang_RDSNet_AAAI_2020} typically aim at faster inference speed by avoiding proposal generation and feature pooling strategies. However, most single-stage approaches are generally inferior in accuracy compared to their two-stage counterparts. Recently, YOLACT~\cite{Bolya_YOLACT_ICCV_2019} obtains an optimal tradeoff between accuracy and speed by predicting a dictionary of category-independent maps (basis masks) for an image and a single set of instance-specific coefficients. Despite its real-time capabilities, YOLACT achieves inferior accuracy compared to two-stage methods.  Different to YOLACT, which has a single set of coefficients for each bounding-box (Fig.~\ref{fig:progress}(c)), our novel SP module aims at preserving spatial information within a bounding-box. The SP module generates multiple sets of spatial coefficients that splits mask prediction into different sub-regions in a bounding-box (Fig.~\ref{fig:progress}(d)). Further, SP module contains a feature alignment scheme that improves feature representation by aligning the predicted instance mask with detected bounding-box. Our SP module is different to feature pooling strategies, such as PSRoI~\cite{Li_FCIS_CVPR_2017} in several ways. Instead of pooling features into a fixed size ($k \times k$), we perform a simple linear combination between spatial coefficients and basis masks without any feature resizing operation. This preservation of feature resolution is especially suitable for large objects. PSRoI pooling (Fig.~\ref{fig:progress}(b)) generates feature maps of $2(c + 1) \times k \times k$ channels, where $k$ is the pooled feature size and $c$ is the number of classes. In practice, such a pooling operation is memory expensive (7938 channels for $k=7$ and $c=80$). Instead, our design is memory efficient since the basis masks are of only $32$ channels for whole image and the spatial coefficients are a $32$ dimensional vector for each sub-region of a bounding-box (Fig.~\ref{fig:progress}(d)). Further, compared to contemporary work \cite{Chen_BlendMask_CVPR_2020} using RoIpool based feature maps, our approach utilizes fewer coefficients on original basis mask. Moreover, our SipMask can be adapted for real-time single-stage video instance segmentation.

\section{Method}

\begin{figure*}[t!]
\centering
\includegraphics[width=0.9\textwidth]{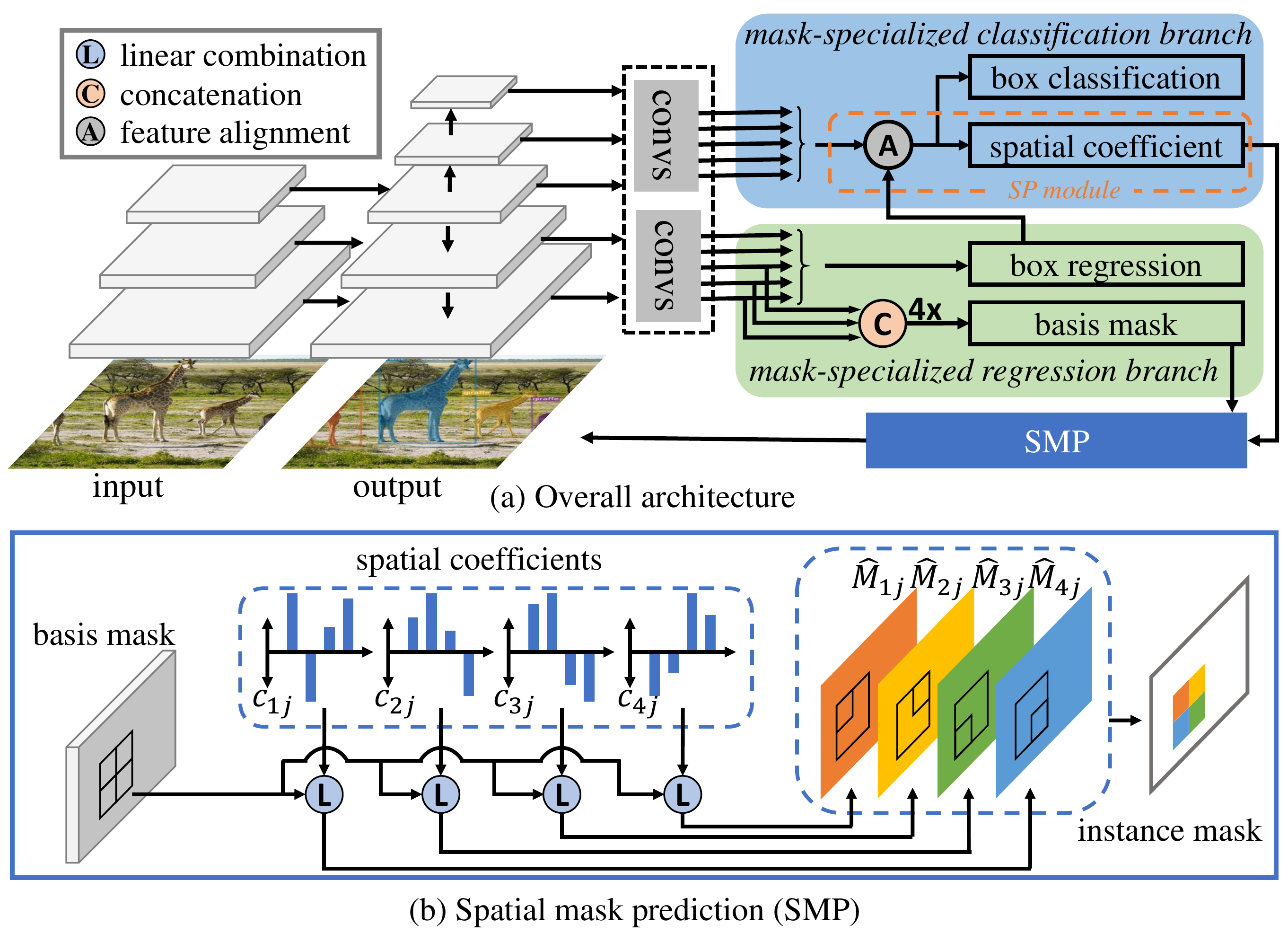}
\caption{(a) Overall architecture of our SipMask comprising fully convolutional mask-specialized classification (Sec.~\ref{sec:sp}) and regression (Sec.~\ref{sec:mbm}) branches. The focus of our design is the introduction of a novel spatial preservation (SP) module in the  mask-specialized classification branch. The SP module performs two-tasks: feature alignment and spatial coefficients generation. In our approach, a separate set of spatial coefficients are generated for each predicted bounding-box. These spatial coefficients are designed to preserve the spatial information within an object instance, thereby enabling improved  delineation  of spatially adjacent objects. The mask-specialized regression branch predicts both bounding-box offsets and a set of category-independent basis masks. The basis masks are generated by capturing contextual information from different prediction layers of FPN. (b) Both the basis masks and spatial coefficients along with predicted bounding-box locations are then input to our spatial mask prediction (SMP) module (Sec.~\ref{sec:smp}) for predicting the final instance mask.
} 
\label{fig:fig_method}
\end{figure*} 

\noindent{\textbf{Overall Architecture:}} Fig.~\ref{fig:fig_method}(a) shows the overall architecture of our single-stage anchor-free method, SipMask, named for its instance-specific spatial information preservation characteristic. Our architecture is built on FCOS detection method~\cite{Tian_FCOS_ICCV_2019}, due to its flexible anchor-free design. In the proposed architecture, we replace the standard classification and regression in FCOS with our mask-specialized regression and classification branches. Both mask-specialized classification and regression branches are fully convolutional. Our mask-specialized classification branch predicts the classification scores of detected bounding-boxes and generates instance-specific spatial coefficients for instance mask prediction. The focus of our design is the introduction of a novel spatial preservation (SP) module, within the mask-specialized classification branch, to obtain improved mask predictions. Our SP module further enables better delineation of spatially adjacent objects. The SP module first performs feature alignment by using the final regressed bounding-box locations. The resulting aligned features are then utilized for both box classification and generating spatial coefficients required for mask prediction. The spatial coefficients are introduced to preserve spatial information within an object bounding-box. In our framework, we divide the bounding-box into $k \times k$ sub-regions and compute a separate set of spatial coefficients for each sub-region. Our mask-specialized regression branch generates both bounding-box offsets for each instance and a set of category-independent maps, termed as basis masks, for an image. Our basis masks are constructed by capturing the contextual information from different prediction layers of FPN.

The spatial coefficients predicted for each of $k \times k$ sub-regions within a bounding-box along with image-specific basis masks are utilized in our spatial mask prediction (SMP) module (Fig.~\ref{fig:fig_method}(b)). Our SMP generates separate map predictions for respective regions within the bounding-box. Consequently, these separate map predictions are combined to obtain final instance mask prediction. 

\subsection{Spatial Preservation Module} \label{sec:sp}
Besides box classification, our mask-specialized classification branch comprises a novel spatial preservation (SP) module. Our SP module performs two tasks: spatial coefficients generation and feature alignment. 
The spatial coefficients are introduced to improve mask prediction by preserving spatial information  within a bounding-box. 
Our feature alignment scheme aims at improving the feature representation for both box classification and spatial coefficients generation.\\
\noindent{\textbf{Spatial Coefficients Generation:}}
As discussed earlier, the recently introduced YOLACT~\cite{Bolya_YOLACT_ICCV_2019} utilizes a single set of coefficients to predict the whole mask of an object, leading to the loss of spatial information within a bounding-box.
To address this issue, we propose a simple but effective approach that splits mask prediction into multiple sub-mask predictions. We divide the spatial regions within a predicted bounding-box into $k \times k$ sub-regions. Instead of predicting a \textit{\textit{single} set of coefficients}  for the whole bounding-box $j$, we predict a \textit{separate set of spatial coefficients} ${c}_{ij} \in R^{m}$ for each of its sub-region $i$. Fig.~\ref{fig:fig_method}(b) shows an example where a bounding-box is divided into $2\times2$ sub-regions (four quadrants, \ie, top-left, top-right, bottom-left and bottom-right). In practice, we observe that $k = 2$ provides an optimal tradeoff between speed and accuracy. Note that our spatial coefficients utilize improved features obtained though a feature alignment operation described next. 

\noindent{\textbf{Feature Alignment Scheme:}} Generally, convolutional layer operates  on a rectangular grid (\eg, $3\times 3$ kernel). Thus, the extracted features for classification and coefficients generation may fail to align with the features of regressed bounding-box. Our feature alignment scheme addresses this issue by  aligning  the features with regressed box location, resulting in an improved feature representation. For feature alignment, we introduce a deformable convolutional layer \cite{Dai_DCN_ICCV_2017,Yang_DenseRepPoints_ECCV_2020,Cao_HSD_ICCV_2019} in our mask-specialized classification branch. The input to the deformable convolutional layer are the regression offsets to left, right, top, and bottom corners of ground-truth bounding-box  obtained from mask-specialized regression branch (Sec.~\ref{sec:mbm}). These offsets are utilized to estimate the kernel offset $\Delta p_r$ that augments the regular sampling grid $G$ in  the deformable convolution operator, resulting in an aligned feature $y(p_0)$ at position $p_0$, as follows:
\begin{equation}
y(p_0)=\sum_{i\in G}  w_r \cdot  x(p_0+p_r+\Delta  p_r),
\end{equation}
where $x$ is the input feature, and  ${p_r}$ is the original position of convolutional weight ${w_r}$ in  $G$. Different to \cite{Yang_DenseRepPoints_ECCV_2020,Yang_RepPoints_ICCV_2019} that aim to learn accurate geometric localization,  our approach aims  to generate better features for box classification and coefficient generation. Next, we describe mask-specialized regression branch.

\subsection{Mask-specialized Regression Branch}\label{sec:mbm}
Our mask-specialized regression branch performs box regression and generates a set of category-independent basis masks for an image. Note that YOLACT utilizes a single FPN prediction layer to generate the basis masks. Instead, the basis masks in our SipMask are generated by exploiting the multi-layer information from different prediction layers of  FPN. The incorporation of multi-layer information  helps to obtain a continuous mask (especially on large objects) and remove background clutter. Further, it helps in scenarios, such as partial occlusion and large-scale variation.
Here, objects of various sizes are predicted at different prediction layers of the FPN (\ie, $P3-P7$).  To capture multi-layer  information, the features from the $P3-P5$ layers of the FPN are utilized to generate basis masks.
Note that $P6$ and $P7$ are excluded for basis mask generation to reduce the computational cost.
The outputs from  $P4$ and $P5$ are first upsampled to the resolution of  $P3$ using bilinear interpolation. The resulting features from all three prediction layers ($P3-P5$) are  concatenated, followed by  a $3\times3$ convolution to generate feature maps with $m$ channels. Finally, these feature maps are upsampled four times by using bilinear interpolation,
resulting in  $m$ basis masks, each having a spatial resolution of $h\times w$. Both the spatial coefficients (Sec.~\ref{sec:sp}) and basis masks are utilized in our spatial mask prediction (SMP) module for final instance mask prediction.

\subsection{Spatial Mask Prediction Module}\label{sec:smp}
\begin{figure}[t!]
\centering
\includegraphics[width=\linewidth]{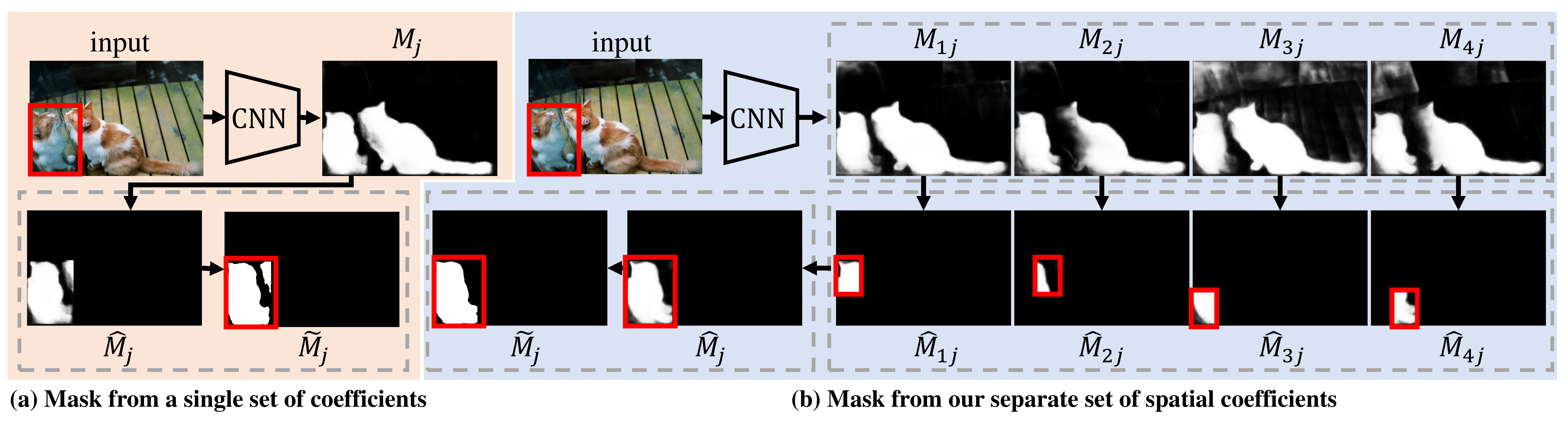} 
\caption{A visual comparison between mask generation using (a) a single set of coefficients, as in YOLACT and (b) our SipMask. For simplicity, only one detected `cat' instance and its corresponding mask generation procedure is shown here. A linear combination of single set of coefficients and basis masks leads to one map ${M}_{j}$. Then, the map ${M}_{j}$ is pruned followed by thresholding to produce the final mask $\tilde{M}_j$. Instead, our SipMask generates a separate set of spatial coefficients for each sub-region (quadrant for $k = 2$) within a bounding-box. As a result, a separate set of spatial map ${M}_{ij}$ is obtained for each quadrant $i$ in the bounding-box $j$. Afterwards, these spatial maps are first pruned and then integrated (a simple addition) followed by thresholding to obtain final mask $\tilde{M}_j$. Our SipMask is able to reduce the influence of the adjacent object (`cat') instance, resulting in improved mask prediction.}
\label{fig:step} 
\end{figure}

Given an input image, our spatial mask prediction (SMP) module takes the predicted bounding-boxes, basis masks and  spatial coefficients as inputs and predicts the final instance mask. 
Let $B \in R^{ h\times w \times m}$ represent $m$ predicted basis masks for the whole image,  $p$ be the number of predicted boxes, and 
${C}_{i}$ be a $m\times p$ matrix that indicates the spatial coefficients at the $i^{th}$ sub-region (quadrant for $k = 2$) of all $p$  predicted bounding-boxes.  Note that the column $j$  of ${C}_{i}$ (\ie, $c_{ij}\in R^m$) indicates the spatial coefficients for the bounding-box $j$ (Sec.~\ref{sec:sp}). 
We perform a simple matrix multiplication  between ${C}_{i}$ and $B$ to obtain $p$  maps  corresponding to the $i^{th}$ quadrant of all bounding-boxes as follows.  
\begin{equation}
M_{i} = \sigma(B\times C_{i}) \; \; \; \forall i\in[1,4], 
\end{equation}
where $\sigma$ is sigmoid  normalization and $M_{i}\in R^{h\times w\times p}$ are the maps generated for the $i^{th}$  quadrant of all $p$ bounding-boxes. Fig.~\ref{fig:step}(b) shows the procedure to obtain final mask of an instance $j$. 
Let $M_{ij}\in R^{h\times w}$ be the map generated for the $i^{th}$ quadrant of a bounding-box $j$. Then, the response values of $M_{ij}$ outside the $i^{th}$ quadrant of the box $j$  are set as zero for generating a pruned map $\hat{M}_{ij}$. To obtain the instance map $\hat{M}_j$ of a bounding-box $j$, we  perform a simple addition of its pruned maps obtained  from all four quadrants, \ie, $\hat{M}_j=\sum_{i=1}^{4} \hat{M}_{ij}$. Finally, the instance map at the  predicted bounding-box region is binarized with a fixed threshold to obtain final mask $\tilde{M}_j$ of instance $j$.

Fig.~\ref{fig:step} shows a visual comparison of a single set of coefficients based mask prediction, as in YOLACT, with our separate set of spatial coefficients (for each sub-region) based mask prediction. The top-left pixels of an adjacent `cat' instance are appearing inside the top-right quadrant of the detected `cat' instance bounding-box (in red).
In Fig.~\ref{fig:step}(a), a linear combination of a single set of instance-specific coefficients  and image-level basis masks is used to obtain a map $M_{j}$. The response values of the map ${M}_{j}$ outside the box $j$ are assigned with zero to produce a pruned mask $\hat{M}_{j}$, followed by thresholding to obtain the final mask $\tilde{M}_j$. 
Instead, our SipMask (Fig.~\ref{fig:step}(b)) generates a separate set of instance-specific spatial coefficients for each sub-region $i$ within a bounding-box $j$.  
By separating the mask predictions to different sub-regions of a box, our SipMask reduces the influence of adjacent (overlapping) object instance in final mask prediction.

\subsection{Loss Function}
\label{sec:loss}
The overall loss function of our framework contains loss terms corresponding to bounding-box detection (classification and regression) and mask generation. For box classification $L_{cls}$ and box regression $L_{reg}$, we utilize focal loss and IoU loss, respectively, as in~\cite{Tian_FCOS_ICCV_2019}. For mask generation, we introduce a novel mask alignment weighting loss $L_{mask}$ that better correlate mask predictions with high quality bounding-box detections. Different to YOLACT that utilizes a standard  pixel-wise binary cross entropy (BCE) loss during training, our $L_{mask}$ improves the BCE loss with a mask alignment weighting scheme that assigns higher weights to the masks $\tilde{M}_j$  obtained from high quality bounding-box detections.\\
\noindent{\textbf{Mask Alignment Weighting:}} In our mask alignment weighting, we first compute the  overlap $o_j$ between a predicted bounding-box $j$  and the corresponding ground-truth. The weighting factor  $\alpha_j$ is then obtained by multiplying the overlap $o_j$ and the classification score $s_j$ of the bounding-box $j$. Here, a higher  $\alpha_j$ indicates good quality bounding-box detections. Consequently, $\alpha_j$ is used to weight the mask loss $ l^j$ of the instance $j$, leading to  $L_{mask} = \frac{1}{N}\sum_{j} l^j\times \alpha_j$. Here, $N$ is the number of bounding-boxes.   Our weighting strategy encourages the network to predict a high quality instance mask for a high quality bounding-box detections. The proposed mask alignment weighting loss $L_{mask}$ is utilized along with loss terms corresponding to bounding-box detection (classification and regression) in our overall loss function: $L=L_{reg}+L_{cls}+L_{mask}$. 

\subsection{Single-stage Video Instance Segmentation}
\label{sec:vis}
In addition to still image instance segmentation, we investigate our single-stage SipMask for the problem of real-time video instance segmentation. In video instance segmentation, the aim is to simultaneously detect, segment, and track instances in videos. 

To perform real-time single-stage video instance segmentation, we simply extend our SipMask by introducing an additional fully-convolutional branch in parallel to mask-specialized classification and regression branches for instance tracking. The fully-convolutional branch consists of two convolutional layers. After that, the output feature maps of different layers in this branch are fused to obtain the tracking feature maps, similar to basis mask generation in our mask-specialized regression branch. Different from the state-of-the-art MaskTrack R-CNN~\cite{Yang_VIS_ICCV_2019} that utilizes RoIAlign and fully-connected operations, our SipMask extracts a tracking feature vector from the tracking feature maps at the bounding-box center to represent each instance. The metric for matching the instances between different frames is similar to MaskTrack R-CNN. Our SipMask is very simple, efficient and achieves favourable performance for video instance segmentation (Sec.~\ref{exp:vis}).

\section{Experiments}
\subsection{Dataset and Implementation Details}
\noindent \textbf{Dataset:} We conduct experiments on COCO dataset~\cite{Lin_COCO_ECCV_2014}, where the \texttt{trainval} set has about 115$k$ images, the \texttt{minival} set has 5$k$ images, and the \texttt{test-dev} set has about 20$k$ images. We perform training on \texttt{trainval} set and  present  state-of-the-art comparison on \texttt{test-dev} set and the ablations on \texttt{minival} set.

\noindent \textbf{Implementation Details:} We adopt ResNet~\cite{He_ResNet_CVPR_2016} (ResNet50/ResNet101) with FPN pre-trained on ImageNet~\cite{Russakovsky_ImageNet_IJCV_2015} as the backbone. Our method is trained eight GPUs with SGD for optimization. During training, the initial learning rate is set to 0.01. When conducting ablation study, we use 
a 1$\times$ training scheme at single scale to reduce training time.   For a fair comparison with the  state-of-the-art single-stage  methods \cite{Chen_TensorMask_ICCV_2019,Bolya_YOLACT_ICCV_2019}, we follow  the 6$\times$, multi-scale  training scheme. During inference  we select  top 100 bounding-boxes  with highest  classification scores, after NMS. 
For these bounding-boxes, a simple linear combination between the predicted spatial coefficients and  basis masks are used  to obtain instance  masks.

\begin{table*}[t!]
\renewcommand{\arraystretch}{1.0}
\centering
\caption{State-of-the-art instance segmentation comparison in terms of accuracy (mask AP) and speed (inference time) on COCO \texttt{test-dev} set. All results are based on single-scale test and speeds are reported on a single Titan Xp GPU (except TensorMask and RDSNet that are reported on Tesla V100). 
When using the same large input size ($\sim 1333 \times 800 $) and backbone, our SipMask outperforms all existing single-stage methods in terms of accuracy. Further, our SipMask obtains a four-fold speedup over the TensorMask. When using a similar small input size ($\sim 550 \times 550 $), our SipMask++ achieves superior performance while operating at comparable speed, compared to the YOLACT++. In terms of real-time capabilities, our SipMask consistently improves the mask accuracy without any significant reduction in speed, compared to the YOLACT.  
} 
\resizebox{\textwidth}{!}{
\begin{tabular}{|l|l|c|c|c|cc|ccc|}
\hline
Method       & Backbone      & Input Size  & Time  & AP & AP@0.5 & AP@0.75 & AP$_s$ & AP$_m$ & AP$_l$ \\
\hline \hline
\textbf{Two-Stage:} & & & & & & & & &\\
MNC \cite{Dai_MNC_CVPR_2016}            & ResNet101-C4  & $\sim1333\times 800$  &  -  & 24.6 & 44.3 & 24.8 & 4.7 & 25.9 & 43.6\\
FCIS  \cite{Li_FCIS_CVPR_2017}          & ResNet101-C5   & $\sim1333\times 800$ &  152  & 29.2 & 49.5 & - & 7.1 & 31.3 & 50.0\\
RetinaMask  \cite{Fu_retinamask_arXiv_2019}        & ResNet101-FPN    & $\sim1333\times 800$ &  167  & 34.7 &  55.4 & 36.9 &  14.3 &  36.7 & 50.5 \\
MaskLab  \cite{Chen_MaskLab_CVPR_2018}        & ResNet101    & $\sim1333\times 800$  &  - & 35.4 &  57.4 & 37.4 &  16.9 &  38.3 & 49.2 \\
Mask R-CNN  \cite{He_MaskRCNN_ICCV_2017}          & ResNet101-FPN    & $\sim1333\times 800$  &  \textbf{116} & 35.7 & 58.0 & 37.8 & 15.5 & 38.1 & 52.4\\
Mask R-CNN*  \cite{He_MaskRCNN_ICCV_2017}          & ResNet101-FPN    & $\sim1333\times 800$  &  116 & 38.3 & 61.2 & 40.8 & 18.2 & 40.6 & 54.1\\
PANet \cite{Liu_PANet_CVPR_2018}   & ResNet50-FPN    & $\sim1333\times 800$  & 212 &  36.6&  58.0 &  39.3 & 16.3 &  38.1 &  53.1\\
MS R-CNN  \cite{Huang_MSRCNN_CVPR_2019}         & ResNet101-FPN    & $\sim1333\times 800$ &  117  & 38.3 & 58.8 & 41.5 & 17.8 &  40.4 & \textbf{54.4}\\
HTC  \cite{Chen_HTC_CVPR_2019}         & ResNet101-FPN    & $\sim1333\times 800$ & 417 &  39.7 &  \textbf{61.8} &  43.1 &  21.0 & 42.2 &  53.5\\
D2Det \cite{Cao_D2Det_CVPR_2020}         & ResNet101-FPN    & $\sim1333\times 800$ & 168 &  \textbf{40.2} &  61.5 &  \textbf{43.7} &  \textbf{21.7} & \textbf{43.0} &  54.0\\
\hline \hline
\textbf{Single-Stage:}  \footnotesize{Large input size}& & & & & & & & &\\
PolarMask  \cite{Xie_PolarMask_arXiv_2019}       & ResNet101-FPN    & $\sim1333\times 800$ & - & 30.4  & 51.9 & 31.0 & 13.4 & 32.4 & 42.8 \\
RDSNet   \cite{Wang_RDSNet_AAAI_2020}      & ResNet101-FPN    & $\sim1333\times 800$ & 113 &  36.4 & 57.9 & 39.0 & 16.4 & 39.5 & 51.6\\
TensorMask  \cite{Chen_TensorMask_ICCV_2019}       & ResNet101-FPN  &  $\sim1333\times 800$  &  380 & 37.1 & 59.3 & 39.4 & 17.1 & 39.1 & 51.6\\
\textbf{Our SipMask}         & ResNet101-FPN   & $\sim1333\times 800$  &  \textbf{89} & \textbf{38.1} & \textbf{60.2}  & \textbf{40.8} & \textbf{17.8} & \textbf{40.8} & \textbf{54.3}\\
\hline \hline
\textbf{Single-Stage:} \footnotesize{Small input size}& & & & & & & & &\\
YOLACT++  \cite{Bolya_YOLACT++_arXiv_2020}      & ResNet101-Deform &     $550\times550$& \textbf{37} &  34.6 & 53.8 & 36.9 & \textbf{11.9} & 36.8 & 55.1\\
\textbf{Our SipMask++}         & ResNet101-Deform   & $544\times544$ & \textbf{37}  & \textbf{35.4} & \textbf{55.6} & \textbf{37.6} & 11.2 & \textbf{38.3} & \textbf{56.8}\\
\hline
\hline
\textbf{Real-Time:} & & & & & & & & & \\
YOLACT   \cite{Bolya_YOLACT_ICCV_2019}       & ResNet50-FPN &     $550\times550$& \textbf{22} &  28.2  & 46.6  & 29.2  & 9.2  & 29.3  & 44.8 \\
\textbf{Our SipMask}        & ResNet50-FPN   & $544\times 544$ &  24  & 31.2 & 51.9 & 32.3 & 9.2 & 33.6 & 49.8\\
\hline
YOLACT   \cite{Bolya_YOLACT_ICCV_2019}       & ResNet101-FPN &     $550\times550$& 30 & 29.8  & 48.5 & 31.2 & \textbf{9.9} & 31.3 & 47.7 \\
\textbf{Our SipMask}          & ResNet101-FPN   & $544\times 544$ &  32  & \textbf{32.8} & \textbf{53.4}  & \textbf{34.3} & 9.3 & \textbf{35.6} & \textbf{54.0}\\
\hline
\end{tabular}}
\label{tab_stateofart}
\end{table*}

\subsection{State-of-the-art Comparison}
Here, we compare our method with some two-stage~\cite{Dai_MNC_CVPR_2016,Li_FCIS_CVPR_2017,Fu_retinamask_arXiv_2019,Chen_MaskLab_CVPR_2018,He_MaskRCNN_ICCV_2017,Liu_PANet_CVPR_2018,Huang_MSRCNN_CVPR_2019,Chen_HTC_CVPR_2019,Cao_D2Det_CVPR_2020} and single-stage~\cite{Zhou_ExtremeNet_CVPR_2019,Bolya_YOLACT_ICCV_2019,Xie_PolarMask_arXiv_2019,Chen_TensorMask_ICCV_2019} methods on COCO \texttt{test-dev} set. Tab.~\ref{tab_stateofart} shows the comparison in terms of both speed and accuracy. Most existing methods use a larger input image size, typically $\sim 1333  \times 800$ (except YOLACT~\cite{Bolya_YOLACT_ICCV_2019}, which operates on input size of $550 \times 550 $). Among existing two-stage methods, Mask R-CNN~\cite{He_MaskRCNN_ICCV_2017} and PANet~\cite{Liu_PANet_CVPR_2018} achieve overall mask AP scores of 35.7 and 36.6, respectively. The  recently introduced MS R-CNN~\cite{He_MaskRCNN_ICCV_2017} and HTC~\cite{Chen_HTC_CVPR_2019}  obtain mask AP scores of 38.3 and 39.7, respectively. Note that HTC achieves this improved accuracy at the cost of a significant reduction in speed. Further, most two-stage approaches require more than 100 milliseconds (ms) to process an image. 

In case of single-stage methods, PolarMask~\cite{Xie_PolarMask_arXiv_2019} obtains a mask AP of 30.4. RDSNet~\cite{Wang_RDSNet_AAAI_2020} achieves a mask AP score of 36.4. Among these single-stage methods, TensorMask~\cite{Chen_TensorMask_ICCV_2019} obtains the best results with a mask AP score of 37.1. Our SipMask under similar settings (input size and backbone) outperforms TensorMask with an absolute gain of 1.0\%, while obtaining a four-fold speedup. In particular, our SipMask achieves an absolute gain of 2.7\% on the large objects, compared to TensorMask. 

\begin{figure*}[t!]
	\includegraphics[width=\textwidth]{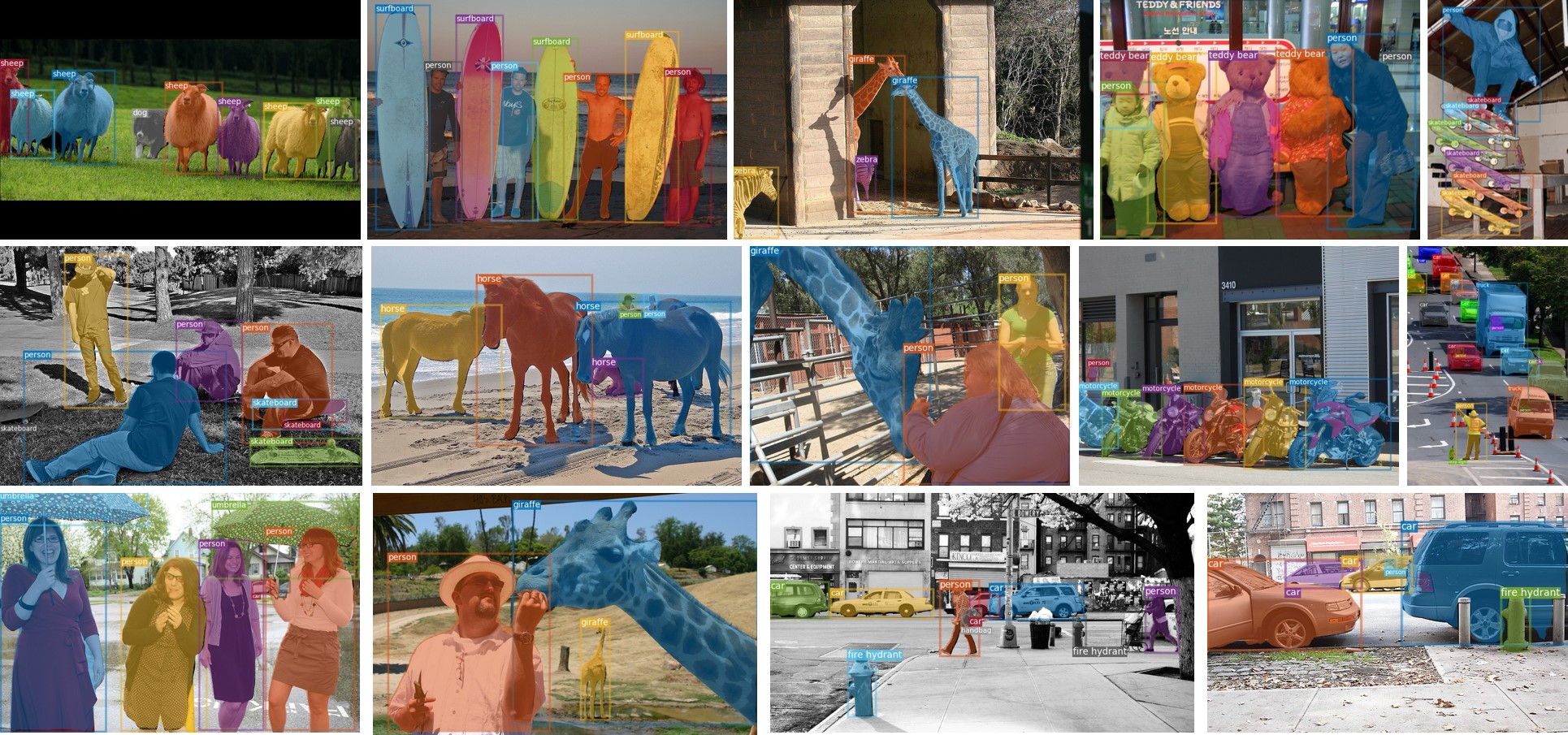}
	\caption{Qualitative results on COCO~\texttt{test-dev} \cite{Lin_COCO_ECCV_2014} (corresponding to our 38.1 mask AP). Each color represents different object instances in an image. Our SipMask generates high quality instance segmentation masks in challenging scenarios.
	}
	\label{fig:qtest}
\end{figure*}

In terms of fast instance segmentation and real-time capabilities, we compare our SipMask with YOLACT~\cite{Bolya_YOLACT_ICCV_2019} when using two different backbone models (ResNet50/ResNet101 FPN). Compared to YOLACT, our SipMask achieves an absolute gain of 3.0\% without any significant reduction in speed (YOLACT: 30 ms vs. SipMask: 32 ms). 
A recent variant of YOLACT, called YOLACT++~\cite{Bolya_YOLACT++_arXiv_2020}, utilizes a deformable backbone (ResNet101-Deform \cite{Zhu_DCNV2_CVPR_2019} with interval 3) and a mask scoring strategy. For a fair comparison, we also integrate the same two ingredients in our SipMask, called as SipMask++. When using a similar input size and same backbone, our SipMask++ achieves improved mask accuracy while operating at the same speed, compared to YOLACT++. Fig.~\ref{fig:qtest} shows example instance segmentation  results of our SipMask on COCO \texttt{test-dev}.

\begin{table}[t]\setlength{\tabcolsep}{4pt}
\parbox{0.475\linewidth}{
\centering
\caption{Impact of progressively integrating (from left to right) different components into the baseline. All our components (SP, CBM and WL) contribute towards achieving improved mask AP.}
\begin{tabular}{|cccc|c|}
\hline
Baseline &  SP  & CBM   & WL  & AP  \\  \hline \hline
\checkmark &    &  & & 31.2\\ 
\checkmark   & \checkmark & &  & 33.4 \\ 
\checkmark  & \checkmark   & \checkmark & & 33.8 \\ 
\checkmark  & \checkmark   & \checkmark & \checkmark & 34.3 \\ 
\hline
\end{tabular}    
\label{tab:integrate}
}
\hfill
\parbox{0.475\linewidth}{
\centering
\caption{Impact of integrating different components individually into the baseline. Our spatial coefficients (SC) obtains the most improvement in accuracy.} 
\begin{tabular}{|ccccc|c|}
\hline
Baseline & SC & FA  & CBM & WL  & AP  \\
\hline
\hline
\checkmark & &  &  &   & 31.2  \\ 
\checkmark & \checkmark ~~&  &  &   & 32.9  \\ 
\checkmark &~~~~ &  \checkmark  &  &   & 31.7  \\ 
\checkmark &~~ &    & \checkmark &   & 31.9  \\ 
\checkmark &~~ &   &  & \checkmark   & 32.0  \\ 
\hline
\end{tabular}
\label{tab:single}
}
\end{table}

\subsection{Ablation study}
We perform an ablation study on COCO \texttt{minival} set with ResNet50-FPN backbone \cite{Lin_FPN_CVPR_2017}. First, we show the impact of progressively  integrating our different components: spatial preservation (SP) module (Sec.~\ref{sec:sp}), contextual basis masks (CBM) obtained by integrating context information from different FPN prediction layers (Sec.~\ref{sec:mbm}), and mask alignment weighting loss (WL) (Sec.~\ref{sec:loss}), to the baseline. Note that our baseline is similar to YOLACT, obtaining the basis masks by using only  high-resolution FPN layer ($P3$) and using a single set of coefficients for mask prediction. The results are presented in Tab.~\ref{tab:integrate}. The baseline achieves a mask AP of 31.2. All our components (SP, CBM and WL) contribute towards achieving improved performance (mask accuracy). In particular, the most improvement in mask accuracy, over the baseline, comes from our SP module. Our final SipMask integrating all contributions obtains an absolute gain of 3.1\% in terms of mask AP, compared to the baseline. 
We also evaluate the impact of adding our different components individually to the baseline. The results are shown in Tab.~\ref{tab:single}. Among these components, the spatial coefficients provides the most improvement in accuracy over the baseline. It is worth mentioning that both the spatial coefficients and feature alignment constitute our spatial preservation (SP) module. These results suggest that each of our components individually contributes towards improving the final performance. 

\begin{figure}[t!]
\centering
\includegraphics[width=\linewidth]{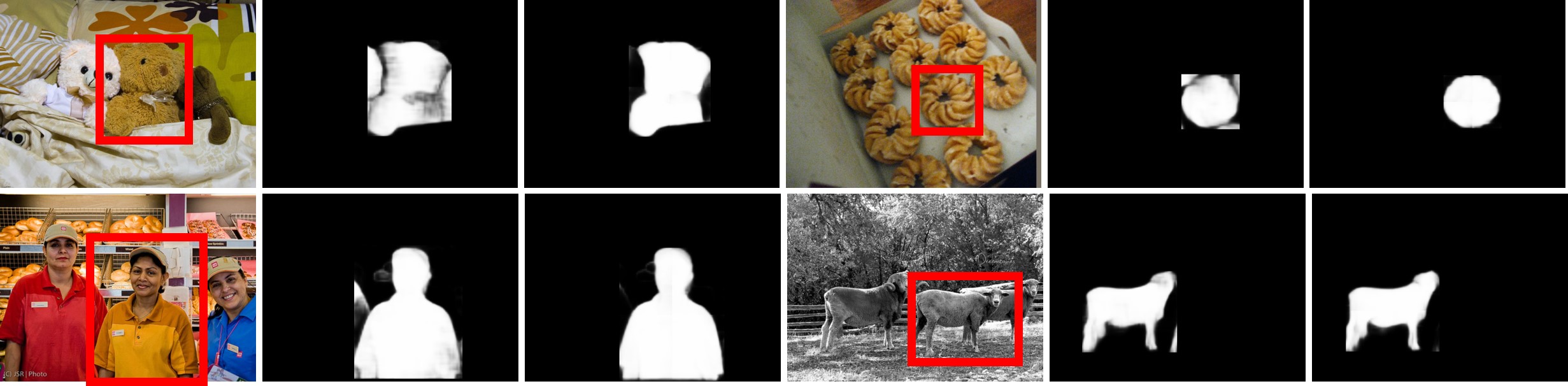}
\caption{Qualitative results highlighting the spatial delineation capabilities of our spatial preservation (SP) module. Input image with a detected bounding-box (red) is shown in column 1 and 4. Mask prediction obtained by the baseline that is based on a single set of coefficients is shown in column 2 and 5. Mask prediction obtained by our approach that is based on a separate set of spatial coefficients in a bounding-box is shown in column 3 and 6. Compared to the baseline, our approach is able to better delineate spatially adjacent object instances, leading to improved mask predictions. } 
\label{fig:vis_Qcompare} 
\end{figure}

Fig.~\ref{fig:vis_Qcompare} shows example results highlighting the spatial delineation capabilities of our spatial preservation (SP) module. We show the input image with the detected bounding-box (red) together with the mask prediction based on a single set of coefficients (baseline) and our mask prediction based on a separate set of spatial coefficients. Our approach is able to provide improved delineation of spatially adjacent instances, leading to superior mask predictions. 

As discussed in Sec.~\ref{sec:sp}, our SP module generates a separate set of spatial coefficients for each sub-region within a bounding-box. Here, we perform a study by varying the number of sub-regions to obtain spatial coefficients. Tab.~\ref{tab:split_comparisons} shows that a large gain in performance is obtained going from $1 \times 1$ to $2 \times 2$. We also observe that the performance tends to marginally increase by further increasing the number of sub-regions. In practice, we found $2 \times 2$ to provide an optimal tradeoff between speed and accuracy. As discussed earlier (Sec.~\ref{sec:loss}), our mask alignment weighting loss re-weights the pixel-level BCE loss using both classification (class scores) and localization (overlap with the ground-truth) information. Here, we analyze the effect of classification (only cls.) and localization (only loc.) on our mask alignment weighting loss in Tab.~\ref{tab:loss_comparison}. It shows that both the classification and localization are useful to re-weight the BCE loss for improved mask prediction.

\begin{table}[t!]
\parbox{0.475\linewidth}{
\centering
\caption{The effect of varying the number of sub-regions to compute spatial coefficients. A separate set of spatial coefficients are generated for each sub-region.}
\begin{tabular}{|c|c|c|c|c|c|c|}
\hline
& $1 \times 1$& $1 \times 2$ & $2 \times 1$ & $2 \times 2 $  & $3 \times 3 $   & $4 \times 4$   \\  \hline \hline 
AP & 31.2 & 32.2 & 32.1 & 32.9 & 33.1 & 33.1 \\ \hline
\end{tabular}    
\label{tab:split_comparisons}
}
\hfill
\parbox{0.475\linewidth}{
\centering
\caption{The effect of classification (class confidences) and localization (ground-truth overlap) scores on our mask alignment weighting loss (cls. + loc.). 
} 
\begin{tabular}{|c|c|c|c|c|}
\hline
& baseline & only cls. & only loc. & cls.+loc.    \\  \hline \hline 
AP & 31.2 & 31.8 & 31.7 & 32.0 \\ \hline
\end{tabular}
\label{tab:loss_comparison}
}
\end{table}

\subsection{Video Instance Segmentation Results}
\label{exp:vis}
\begin{figure*}[t!]
	\includegraphics[width=\textwidth]{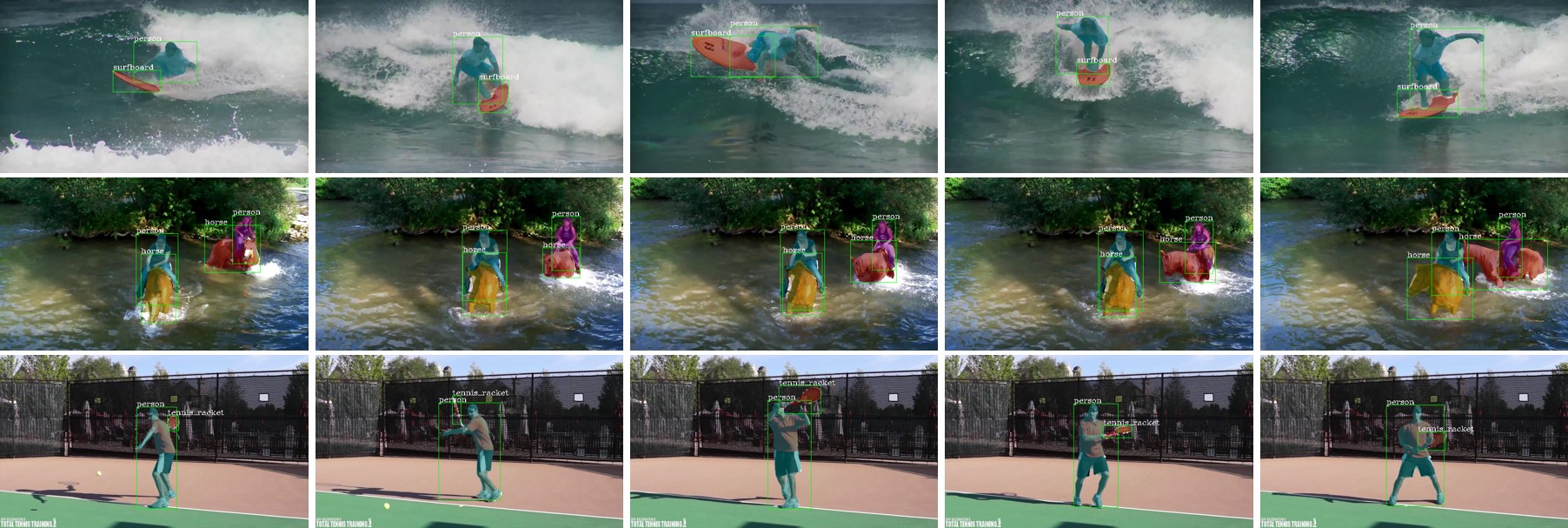}
	\caption{Qualitative results on example frames of different videos from Youtube-VIS validation set \cite{Yang_VIS_ICCV_2019}. The object with same predicted identity has same color.  
	}
	\label{fig:track}
\end{figure*}

In addition to instance segmentation, we present the effectiveness of our SipMask, with the proposed modifications described in Sec.~\ref{sec:vis}, for real-time video instance segmentation. We conduct experiments on the recently introduced large-scale YouTube-VIS dataset \cite{Yang_VIS_ICCV_2019}. The YouTube-VIS dataset contains  2883 videos, 4883 objects, 131$k$ instance masks, and 40 object categories. Tab. \ref{tab:video} shows
the state-of-the-art comparison on the YouTube-VIS validation set. When using the same input size ($640\times360$) and backbone (ResNet50 FPN), our SipMask outperforms the state-of-the-art MaskTrack R-CNN \cite{Yang_VIS_ICCV_2019} with an absolute gain of 2.2\% in terms of mask accuracy (AP). Further, our SipMask achieves impressive mask accuracy while operating at real-time (30 fps) on a Titan Xp. Fig.~\ref{fig:track} shows video instance segmentation results on example frames from the validation set.

\begin{table}[t!]
\centering
\caption{Comparison with state-of-the-art video instance segmentation methods on YouTube-VIS validation set. Results are reported in terms of mask accuracy and recall.}
\resizebox{0.9\linewidth}{!}{
\begin{tabular}{|l|c|c|c|c|c|c|}
\hline
method & category & AP & AP@0.5 & AP@0.75  & AR@1  & AR@10   \\  \hline \hline 
OSMN \cite{Yang_OSMN_CVPR_2018} & mask propagation & 23.4 & 36.5 & 25.7 & 28.9 & 31.1\\
FEELVOS \cite{Voigtlaender_FEELVOS_CVPR_2019} & mask propagation & 26.9 & 42.0 & 29.7 & 29.9 & 33.4\\
\hline
\hline
OSMN \cite{Yang_OSMN_CVPR_2018} & track-by-detect & 27.5 & 45.1 & 29.1 & 28.6 & 31.1\\
MaskTrack R-CNN \cite{Yang_VIS_ICCV_2019} & track-by-detect & 30.3 & 51.1& 32.6 & 31.0 & 35.5\\
\textbf{Our SipMask}  & track-by-detect & 32.5 & 53.0& 33.3 & 33.5 & 38.9\\
\textbf{Our SipMask} \textit{ms-train}  & track-by-detect & \textbf{33.7} & \textbf{54.1}& \textbf{35.8} & \textbf{35.4} & \textbf{40.1}\\
\hline      
\end{tabular}}
\label{tab:video}
\end{table}

\section{Conclusion} 
We introduce a fast single-stage instance segmentation method, SipMask, that aims at preserving spatial information within a bounding-box. A novel light-weight spatial preservation (SP) module is designed to produce a separate set of spatial coefficients by splitting mask prediction of an object into different sub-regions. To better correlate mask prediction with object detection, a feature alignment scheme and a mask alignment weighting loss are further proposed. We also show that our SipMask is easily extended for real-time video instance segmentation. Our comprehensive experiments on COCO dataset show the effectiveness of the proposed contributions, leading to state-of-the-art single-stage instance segmentation performance. With the same instance segmentation framework and just changing the input resolution ($544 \times 544$), our SipMask operates at real-time on a single Titan Xp with a mask accuracy of 32.8 on COCO~\texttt{test-dev}.

This work was supported by National Key R\&D Program (2018AAA0102800) and National Natural Science Foundation (61906131, 61632018) of China.

\clearpage
%
%
\bibliographystyle{splncs04}
\bibliography{egbib}
\end{document}